%% file: main.tex
\documentclass{article}

\usepackage[preprint]{neurips_2026}
\usepackage{wrapfig}

\usepackage[utf8]{inputenc} 
\usepackage[T1]{fontenc}    
\usepackage{hyperref}       
\usepackage{url}            
\usepackage{booktabs}       
\usepackage{amsfonts}       
\usepackage{nicefrac}       
\usepackage{microtype}      
\usepackage{xcolor}         
\usepackage{amsmath}
\usepackage{graphicx}
\usepackage{amssymb}
\usepackage{booktabs}
\usepackage{multirow}
\newcommand{\model}{\textrm{MedKGTab}}
\newcommand{\PrivateDataset}{{\texttt{CAG-Tongue}}}
\newcommand{\PublicDataset}{{\texttt{CAG-Fecal}}}

\usepackage[most]{tcolorbox}
\usepackage{listings}
\usepackage{xcolor}

\definecolor{promptbg}{RGB}{248,248,248}
\definecolor{promptframe}{RGB}{200,200,200}
\definecolor{prompttitle}{RGB}{60,60,60}

\lstdefinelanguage{Prompt}{
  morekeywords={SYSTEM,USER,ASSISTANT,INPUT,OUTPUT,INSTRUCTION,CONTEXT,EXAMPLE,CONSTRAINT},
  sensitive=false,
  morecomment=[l]{\#},
  morestring=[b]"
}

\lstdefinestyle{promptstyle}{
  language=Prompt,
  basicstyle=\ttfamily\small,
  keywordstyle=\bfseries,
  commentstyle=\itshape\color{gray},
  showstringspaces=false,
  breaklines=true,
  columns=fullflexible,
  keepspaces=true,
  frame=none
}

\newtcblisting{promptbox}[2][]{%
  colback=promptbg,
  colframe=promptframe,
  title=#2,
  fonttitle=\bfseries,
  coltitle=prompttitle,
  listing only,
  listing options={style=promptstyle},
  breakable,
  enhanced,
  boxrule=0.6pt,
  arc=2mm,
  left=2mm,
  right=2mm,
  top=1mm,
  bottom=1mm,
  #1
}

\title{Cross-Domain Feature Expansion for Tabular Medical Data via Knowledge Graphs Injection}

\author{%
  Mengying Zhou$^{1}$ \quad Yongjie Yin$^{2}$ \quad Haoyan Xin$^{3}$ \quad Guoping Liu$^{4}$ \quad Yang Chen$^{2}$ \\
  $^{1}$School of Computing and Artificial Intelligence, Shanghai University of Finance and Economics \\
  $^{2}$College of Computer Science and Artificial Intelligence, Fudan University \\
  $^{3}$Independent Researcher \\
  $^{4}$School of Basic Medical Sciences, Shanghai University of Traditional Chinese Medicine and Pharmacology \\
}

\begin{document}

\maketitle
\input{0-abstract}
\input{1-Introduction}
\input{2-Related_Work}
\input{3-Methodology}
\input{4-Experiments}
\input{5-Conclusion}

\bibliographystyle{plain}
\bibliography{refs}

\newpage
\appendix
\input{6-Appendix}


\end{document}

%% file: 0-abstract.tex
\begin{abstract}
\label{sec:abstract}
Acquiring comprehensive cross-domain biomedical profiles is often costly and time-consuming, resulting in severe data scarcity in medical research. To address this challenge, we propose MedKGTab, a knowledge-injected framework specifically engineered for cross-domain feature expansion in tabular medical data. MedKGTab seeks to infer uncollected biomedical features from available ones by exploiting their inherent statistical dependencies and established medical correlations.  
By employing a row-column dual-attention mechanism, MedKGTab operates directly on raw structured tabular data, inherently capturing exact numerical distributions without the structural loss caused by tokenization. Crucially, MedKGTab integrates data-driven statistical priors with the SPOKE biomedical knowledge graph, achieving an optimal synergy between the data and knowledge channels. Within this synergy, the representations derived from the data channel are modulated by the injected biomedical knowledge, ensuring the final generated data are grounded in empirical medical research.
Experimental results demonstrate that MedKGTab achieves high data fidelity and realistic data representation in cross-domain feature expansion. It outperforms both SOTA medical large models (e.g., Baichuan M3-plus) and specialized tabular models designed for medical data generation. 
Furthermore, MedKGTab consistently delivers superior performance across various data generation scenarios, whether inferring missing features within the same dataset or generalizing across different medical cohorts.

\end{abstract}

%% file: 1-Introduction.tex
\section{Introduction}
\label{sec:introduction}

\begin{figure}
    \centering
    \includegraphics[width=1\linewidth]{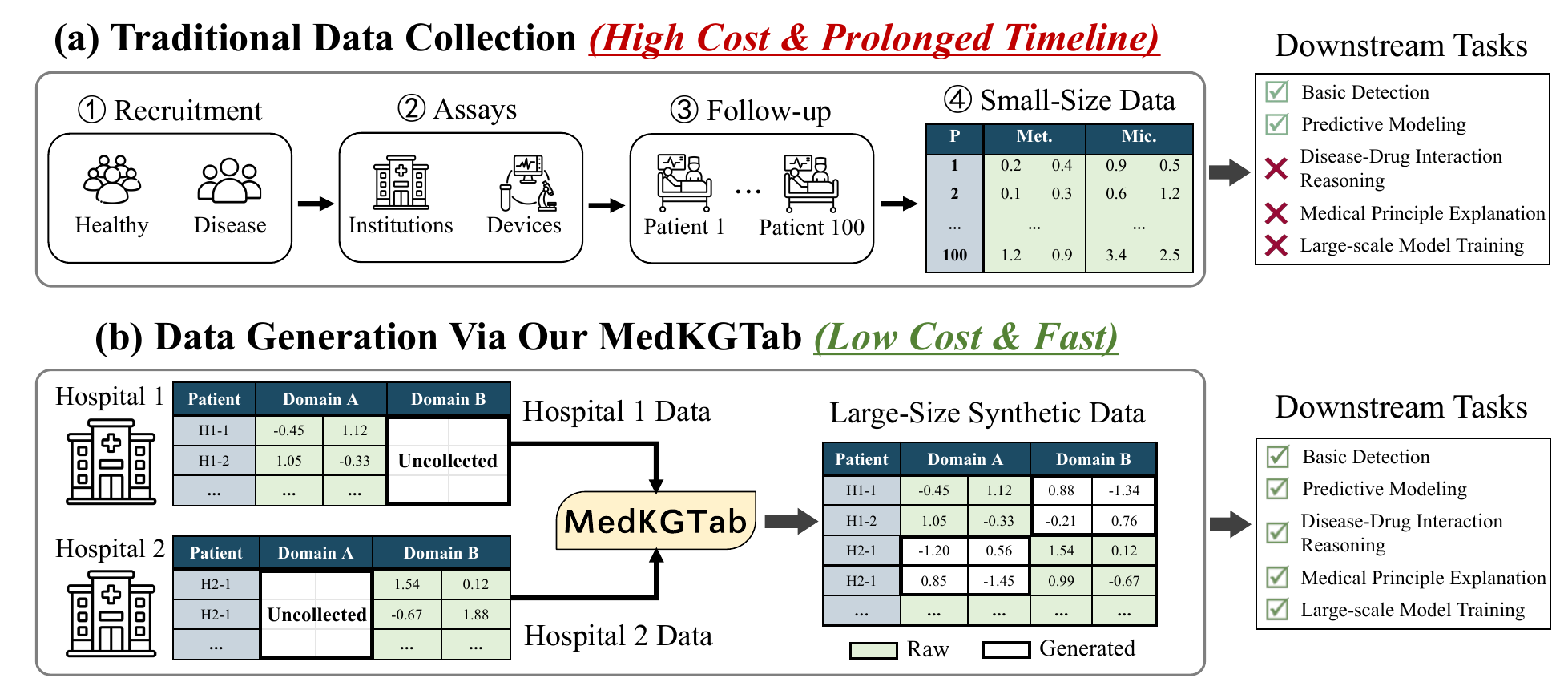}
    \caption{Comparison between the traditional data collection workflow and our knowledge-injected cross-domain feature expansion workflow.}
    \label{fig:workflow}
\end{figure}

Biomedical profiles are essential for medical research since they provide important information for disease characterization, diagnosis, and treatment response analysis. However, acquiring comprehensive cross-domain profiles is prohibitively costly and time-consuming, resulting in severe data scarcity, as depicted in Fig.~\ref{fig:workflow}(a). Cross-domain feature expansion in tabular medical data offers a promising generative solution to infer uncollected features from available ones by exploiting their inherent dependencies~\cite{sun2025dataefficient,wang2026eegdiffuser}. This generative workflow (Fig.~\ref{fig:workflow}(b)) can rapidly and cost-effectively synthesize a large volume of high-fidelity biomedical profiles, which enables broader downstream applications, ranging from basic disease predictive modeling to complex tasks like disease-drug interaction reasoning. 
However, achieving reliable feature expansion across diverse biological domains remains exceptionally difficult due to small sample sizes, severe data sparsity, and complex underlying biological mechanisms ~\cite{athieniti2023guide,mani2025genomics,tarazona2021challenges}.

Existing tabular data generation methods struggle to address these challenges, as they are primarily designed for intra-domain synthesis~\cite{chen2022danets, gorishniy2021revisiting,ye2025revisiting}. To model complex cross-domain dependencies, recent efforts have explored injecting external knowledge graphs (KGs) or developing specialized medical large language models (LLMs), but both paradigms face inherent limitations.
KG injection in tabular data generation models often serves merely as a superficial structural constraint. The impact of this injected knowledge is easily diluted, and its strength lacks flexibility across varying tasks~\cite{choi2017gram}. Meanwhile, although LLMs natively possess vast internal knowledge, they are fundamentally ill-suited for numerical tabular data. By relying on tokenization, LLMs fail to preserve the structural relationships between biological entities~\cite{grinsztajn2022tree}, leading to severe loss of structural information and an inability to capture authentic numerical distributions. Furthermore, the computational costs associated with updating model parameters in LLM-based approaches remain prohibitively high~\cite{li2021prefix, wang2025data}.

To overcome the limitations of both diluted KG injection and structure-agnostic LLMs, we propose MedKGTab, a knowledge-injected framework specifically designed for cross-domain medical feature expansion. By employing a row-column dual-attention mechanism, MedKGTab operates directly on raw structured tabular data, inherently capturing exact numerical distributions without tokenization. More importantly, MedKGTab integrates data-driven statistical priors with the SPOKE biomedical knowledge graph. Unlike previous methods, which easily dilute knowledge, MedKGTab achieves optimal synergy between the data and knowledge channels. By injecting feature-level graph priors into the attention module as a target-specific bias, it explicitly models inter-domain biomedical associations. This mechanism ensures the model receives vital biomedical guidance without overriding its intrinsic data representations, successfully preserving both tabular and graph structural information. In summary, our main contributions are as follows:

\begin{itemize}
    \item We design a lightweight, controllable knowledge injection mechanism that overcomes tabular-graph misalignments. It effectively injects biomedical priors without the high computational cost of updating pre-trained parameters.
    \item We propose MedKGTab, a framework that optimally synergizes data and knowledge channels to provide essential biomedical guidance without overriding intrinsic data representations.
    \item Extensive experiments demonstrate MedKGTab's superior data fidelity and realistic data representation. By capturing numerical distributions and preserving biomedical relationships, it outperforms classical models and specialized medical LLMs (e.g., Baichuan M3-plus) across both intra- and cross-cohort settings.
\end{itemize}

%% file: 2-Related_Work.tex
\section{Related Work}
\label{sec:related_work}

\subsection{Tabular Data Generation}

Synthetic tabular data generation has been widely studied for data augmentation, privacy preservation, and learning under limited supervision. Early deep generative models, such as CTGAN, TVAE~\cite{xu2019modeling}, and TabDDPM~\cite{kotelnikov2023tabddpm}, effectively model mixed numerical and categorical features within a unified framework. 
However, they primarily focus on within-domain generation and struggle with cross-domain medical features. To improve the generalization of tabular models, recent studies have explored pretrained and transferable modeling methods. 
TabPFN~\cite{hollmann2025tabpfn} learns transferable priors through pretraining on more than 100 million synthetic tabular tasks, enabling strong performance in low-data settings. 
FT-Transformer~\cite{gorishniy2021revisiting} and TabR~\cite{gorishniy2024tabr} further improve transferability through stronger architectures or retrieval-based mechanisms. 
MediTab~\cite{wang2024meditab} improves transferability through data consolidation and task alignment across heterogeneous medical tables, while MedTransTab~\cite{chen2025medtranstab} leverages medical LLM. 

Despite these advances, existing methods heavily emphasize predictive performance or statistical fidelity. They do not explicitly incorporate expert biomedical knowledge to reliably govern cross-domain feature dependencies, making them suboptimal for rigorous medical feature expansion.

\subsection{Medical Large Language Models}

Benefiting from pretraining on diverse clinical datasets, medical LLMs have demonstrated strong performance on biomedical reasoning. Representative models such as Med-PaLM 2~\cite{singhal2023medpalm2} and Med-PaLM M~\cite{tu2023medpalmm} demonstrate effectiveness on medical question answering, long-form response generation, and multimodal biomedical understanding. Open medical LLMs, including ChatDoctor~\cite{li2023chatdoctor}, BioMistral~\cite{labrak2024biomistral}, MEDITRON~\cite{chen2023meditron}, and Baichuan-M3~\cite{dou2026baichuanm3}, further improve medical reasoning and complex clinical inference through instruction tuning and domain-specific knowledge injection. 

Although rich in medical knowledge, medical LLMs are designed for unstructured or semi-structured sequences. Applying them directly to tabular feature expansion is severely constrained by limited context windows and a fundamental inability to model complex statistical distributions and quantitative inter-feature correlations.

%% file: 3-Methodology.tex
\section{Methodology}
\label{sec:methodology}

\begin{figure}
    \centering
    \includegraphics[width=\linewidth]{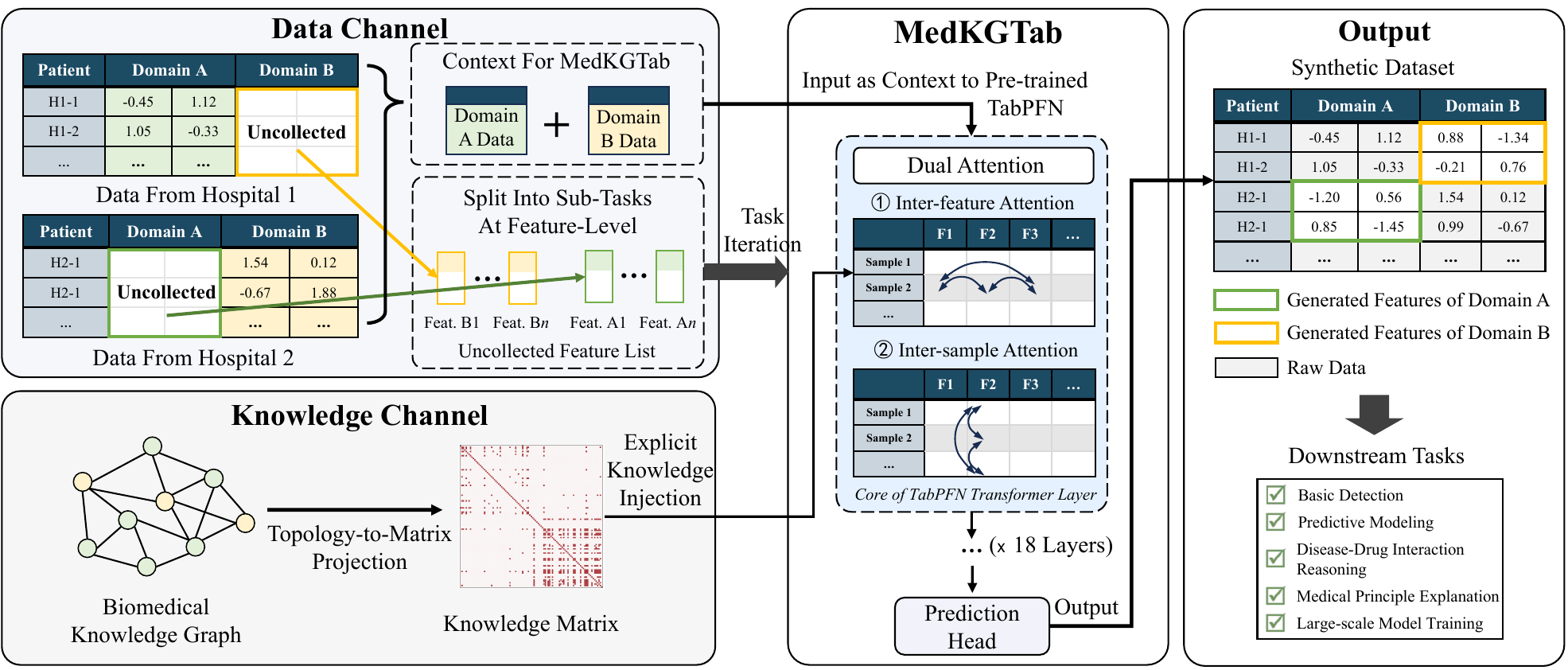}
    \caption{Overall framework of MedKGTab
    }
    \label{fig:framework}
\end{figure}

In the real world, acquiring comprehensive biomedical data is often prohibitively time-consuming and expensive. Consequently, cross-domain feature expansion seeks to infer uncollected target-domain features from readily available source-domain features. Although recent tabular foundation models provide strong transferable priors for small-sample tabular learning, they are prone to learning unstable source-target correlations from limited paired data, often resulting in medically implausible outputs. To address this issue, we propose MedKGTab, a knowledge-injected tabular framework for cross-domain medical feature expansion, as illustrated in Fig.~\ref{fig:framework}. Specifically, we derive a feature-level biomedical prior from an external biomedical knowledge graph and explicitly inject it into the feature attention mechanism of a pre-trained tabular foundation model. Through this synergistic integration, target-domain feature expansion is governed not only by empirical data statistics but also modulated by established biomedical knowledge.

\subsection{Problem Formulation}
For multi-center cross-domain biomedical feature expansion, let the complete global feature space be composed of $D$ distinct clinical variables. In real-world clinical practice, different institutions often record disjoint feature domains. For instance, as illustrated in Fig.~\ref{fig:framework}, the data from Hospital 1 may contain domain A but lack domain B, whereas Hospital 2 exhibits the inverse pattern. 

Suppose we are provided with datasets from multiple distinct cohorts. For the $k$-th cohort comprising $N^{(k)}$ patients, only a specific subset of $d_k$ features is collected, yielding the corresponding data matrix:
$$
\mathcal{D}^{(k)} = \mathbf{X}_{\mathrm{obs}}^{(k)} \in \mathbb{R}^{N^{(k)} \times d_k}.
$$
The complementary feature domains for these patients remain uncollected, denoted as the missing data $\mathbf{X}_{\mathrm{mis}}^{(k)}$.

The overall task is to leverage the aggregated available data across all centers to accurately infer the uncollected feature blocks $\hat{\mathbf{X}}_{\mathrm{mis}}^{(k)}$ for all cohorts $k$, thereby systematically reconstructing a fully complete cross-domain dataset.

\subsection{Data Channel}
\label{sec:data_channel}
As depicted in the top-left panel of Fig.~\ref{fig:framework}, the data channel is designed to structure multi-center datasets into a unified prior context and iterative generation targets. 

First, we aggregate all collected multi-domain data across cohorts to construct a unified context. Next, to cast the multi-target feature expansion as a sequence of tractable inferences, this whole generation task is partitioned into feature-level sub-tasks (e.g., features B1 to B$n$ for Hospital 1). For the $t$-th missing feature, we construct its corresponding input for inference:
$$
\mathcal{D}^{\mathrm{sub}}_t=(\mathbf{X}^{\mathrm{context}}, \mathbf{Y}^{\mathrm{obs}}_{:,t}),
$$
where $\mathbf{X}^{\mathrm{context}}$ is the aggregated context and $\mathbf{Y}^{\mathrm{obs}}_{:,t}$ represents the available data for this feature from other cohorts. MedKGTab then traverses the uncollected feature list to sequentially infer data for each missing feature.

\subsection{Knowledge Channel}
\label{sec:knowledge_channel}

As illustrated in the lower-left panel of Fig.~\ref{fig:framework}, the knowledge channel derives a sample-independent biomedical prior from an external knowledge graph. In this study, we use SPOKE~\cite{morris2023spoke}, a large-scale knowledge graph that integrates diverse entities and relations from curated biomedical resources, providing a structured and reliable basis for constraining feature interactions in medical tabular data.

To align the tabular domain with this graph without semantic ambiguity, we strictly map source and target features to task-relevant node types (e.g., mapping metabolites to \texttt{Compound} nodes and microbial genera to \texttt{Organism} nodes). Each tabular feature is uniquely mapped to at most one corresponding SPOKE node.

Given the inherent sparsity of biomedical networks, relying solely on direct edges is insufficient. Therefore, we perform a topology-to-matrix projection that incorporates closed one-hop neighborhood information, denoted as $\mathcal{N}(\cdot)$. We project this enriched topology into a feature-level knowledge matrix $A \in \mathbb{R}^{(d_x+d_y)\times(d_x+d_y)}$:
$$
A_{ij}=
\begin{cases}
1, & \text{if } i \neq j \text{ and } \mathcal{N}(v_i) \cap \mathcal{N}(v_j) \neq \varnothing,\\
0, & \text{otherwise},
\end{cases}
$$
where $v_i$ and $v_j$ denote the corresponding SPOKE nodes for successfully mapped features $i$ and $j$. Accordingly, $A_{ij}=1$ if the two features share at least one conceptual node in their immediate neighborhoods. For unmapped (out-of-vocabulary) features, their corresponding rows and columns in $A$ remain zero. This projection ensures that $A$ robustly captures feature dependencies fundamentally grounded in shared biomedical topology.

To finalize the prior for model injection, we add self-loops to preserve self-information and apply symmetric normalization:
$$
\tilde{A}=D^{-\frac{1}{2}}(A+I)D^{-\frac{1}{2}},
$$
where $I$ is the identity matrix and $D$ is the degree matrix of $A+I$. The resulting normalized matrix $\tilde{A}$ serves as an interpretable and robust structural prior, mitigating the instability of empirical correlations estimated solely from limited paired tabular data.

\subsection{Knowledge-Injected Dual Attention Inference}
\label{sec:inference}

\subsubsection{Dual Attention Mechanism}
This module serves as the core inference engine, bridging the upstream data and knowledge channels. We employ TabPFN~\cite{hollmann2025tabpfn} as the backbone, leveraging its strong transferable priors cultivated through extensive pre-training, making it exceptionally well-suited for medical scenarios with limited supervision.
Unlike standard sequence-to-sequence Transformers, TabPFN processes tabular context via a dual attention mechanism (as shown in the center panel of Fig.~\ref{fig:framework}): an \textit{inter-feature attention} that captures column-wise variable dependencies, and an \textit{inter-sample attention} that captures row-wise patient relationships.

Specifically, each TabPFN layer consists of a three-stage sequential process: (1) inter-feature attention, (2) inter-sample attention, and (3) feed-forward MLP, formulated as follows:
$$
\begin{aligned}
    \mathbf{H}^{(t, l)}_\mathrm{feat} &= \mathbf{H}^{(t, l)} + \mathrm{Attn}_{\mathrm{feat}}(\mathbf{H}^{(t, l)}), \\
    \mathbf{H}^{(t, l)}_\mathrm{sample} &= \mathbf{H}^{(t, l)}_\mathrm{feat} + \mathrm{Attn}_{\mathrm{sample}}(\mathbf{H}^{(t, l)}_\mathrm{feat}), \\
    \mathbf{H}^{(t, l + 1)} &= \mathbf{H}^{(t, l)}_\mathrm{sample} + \mathrm{MLP}(\mathbf{H}^{(t, l)}_\mathrm{sample}).
\end{aligned}
$$
where $\mathbf{H}^{(t,l)} \in \mathbb{R}^{N \times D \times h}$ denotes the hidden table representation at the $l$-th layer for the $t$-th feature generation sub-task, with $N$ and $D$ representing the sample and feature dimensions of $\mathcal{D}^{\mathrm{sub}}_t$, and $h$ is the hidden dimension.

Our adopted TabPFN v2.5 stacks $L=18$ such layers. The final representation $\mathbf{H}^{(t, L)}$ is then passed to the prediction head to infer the missing values for this specific feature across the required clinical cohorts.

\subsubsection{Explicit Knowledge Injection}
Since our derived biomedical prior knowledge explicitly encodes the relationships between biological variables, we structurally inject this knowledge exclusively into the inter-feature attention. This design choice compels the model to prioritize biologically corroborated feature interactions, while preserving the purely data-driven nature of the inter-sample attention.

To execute this targeted injection, we dynamically construct a task-specific prior for each feature generation. During the task iteration for a given target $y_t$, we extract its corresponding source--source, source--target, and target--target relations from the global prior $\tilde{A}$, forming the sub-matrix $\tilde{A}^{(t)}$:
$$
\tilde{A}^{(t)}=
\begin{bmatrix}
\tilde{A}_{1:d_x,\;1:d_x} & \tilde{A}_{1:d_x,\;d_x+t} \\
\tilde{A}_{d_x+t,\;1:d_x} & \tilde{A}_{d_x+t,\;d_x+t}
\end{bmatrix}
\in \mathbb{R}^{(d_x+1)\times(d_x+1)}.
$$
To match the dimensionality of TabPFN v2.5's internal logits, we apply mean pooling with a window size of $w=3$ to $\tilde{A}^{(t)}$, yielding the final injectable matrix $\tilde{A}^{(t)}_w$.

As TabPFN's default preprocessing disrupts the correspondence between input and original biomedical features, directly injecting the knowledge matrix into TabPFN is infeasible. To preserve this crucial alignment, we disable its internal feature expansion, fingerprint generation, and feature shuffling. These targeted modifications guarantee the backbone's feature indices remain perfectly aligned with our biomedical prior.

Finally, the knowledge matrix $\tilde{A}^{(t)}_w$ is added as a structural bias to the pre-softmax logits of the inter-feature attention. This matrix is shared identically across all attention heads and broadcast across the batch dimension in every attention layer. Formally, the knowledge-injected inter-feature attention is computed as:
$$
\mathrm{Attn}_{\mathrm{feat}}(Q,K,V)=\mathrm{softmax}\left(\frac{QK^\top}{\sqrt{d}}+\alpha \tilde{A}^{(t)}_w\right)V,
$$
where $Q$, $K$, and $V$ represent the query, key, and value matrices of the inter-feature attention, $d$ is the scaling dimension, and $\alpha$ is a tunable hyperparameter controlling the injection strength. By applying $\tilde{A}^{(t)}_w$ as a targeted attention bias, the model is explicitly encouraged to assign larger attention weights to features corroborated by established biomedical knowledge, thereby mitigating the risk of overfitting to spurious correlations in the limited paired data.

\subsection{Output and Downstream}
\label{sec:output_channel}

The output module aggregates the target-wise inferences to construct a complete tabular profile. Once all $d_y$ target features are sequentially generated via the task iterations, we concatenate the predictions along the feature dimension:
$$
\hat{\mathbf{Y}}^{\mathrm{single}} = [\hat{\mathbf{Y}}^{\mathrm{single}}_{:,1}, \hat{\mathbf{Y}}^{\mathrm{single}}_{:,2}, \ldots, \hat{\mathbf{Y}}^{\mathrm{single}}_{:,d_y}].
$$
The final output, $\hat{\mathcal{D}}^{\mathrm{single}} = (\mathbf{X}^{\mathrm{single}}, \hat{\mathbf{Y}}^{\mathrm{single}})$, yields a large-size synthetic dataset containing fully populated cross-domain biomedical profiles, which is ready for diverse downstream clinical analysis.

%% file: 4-Experiments.tex
\section{Experiments}
\label{sec:experiments}
In this section, we evaluate {\model} under intra-cohort and cross-cohort feature expansion tasks, and conduct ablation studies to assess the contribution of knowledge injection.

\subsection{Experimental Setup}

\paragraph{Datasets.}
We evaluate {\model} on two chronic atrophic gastritis~(CAG) datasets, denoted as {\PrivateDataset} and {\PublicDataset}, both comprising paired metabolite and microbiota profiles. {\PrivateDataset} utilizes tongue-coating samples collected from A Hospital (anonymized for double-blind review). After excluding individuals with gastric polyps, hemorrhage, tumors, prior gastrointestinal resection, or special types of gastritis, 114 valid samples remained (78 CAG patients and 36 healthy controls). Each profile includes 432 microbial taxa and 120 metabolites, obtained via 16S rRNA sequencing and UPLC-MS/MS, respectively. 
To evaluate cross-cohort generalization, we further incorporate {\PublicDataset}, a public fecal dataset derived from Gai et al.~\cite{gai2023heptadecanoic}. This dataset contains 33 samples (17 CAG patients and 16 healthy controls), including 147 microbiota and 192 metabolite features.
For all datasets used in this work, we leveraged them solely for secondary analysis and were not involved in their collection or in any clinical experimentation.

\paragraph{Baselines.}
We benchmark {\model} against ten baselines across two categories: dedicated tabular models and LLM-based approaches.
The dedicated tabular baselines include a classical machine learning model, XGBoost~\cite{chen2016xgboost}; deep learning-based tabular models, including FT-Transformer~\cite{gorishniy2021revisiting}, DANets~\cite{chen2022danets}, SwitchTab~\cite{wu2024switchtab}, and modernNCA~\cite{ye2025revisiting}; and a raw tabular foundation model, TabPFN~\cite{hollmann2025tabpfn}.
The LLM-based baselines cover three sub-categories: 
two tabular generation methods, including Curated LLM~\cite{seedat2024curatedllm} and TabGen-ICL~\cite{fang2025tabgen}; a general LLM, DeepSeek v3.2~\cite{liu2025deepseek}; and a SOTA medical LLM, Baichuan M3-plus~\cite{dou2026baichuanm3}, with domain-specialized medical knowledge.

\paragraph{Implementation.}
To accommodate the context window limitations of LLMs, we restrict the input feature space to the top 50 metabolites and top 50 microbial taxa, selected based on feature importance derived from Neural Additive Models (NAM)~\cite{agarwal2021neural}. 

For the intra-cohort setting, we simulate a data-scarce clinical environment by partitioning the patient samples: $40\%$ of the samples are retained as collected data, while the remaining $60\%$ are masked as uncollected data to serve as the generation target. Moreover, we conduct bidirectional inference experiments on intra-cohort settings, treating both microbiota and metabolites as the uncollected target domain in alternating trials.
For the cross-cohort setting (i.e., across different medical institutions), the entirety of {\PrivateDataset} along with one observed domain from {\PublicDataset} is used to infer the remaining uncollected domain in {\PublicDataset}. 

{\model} and all tabular model baselines are implemented in PyTorch and run on a single NVIDIA RTX 4090 GPU (24\,GB memory).
FT-Transformer, DANets, SwitchTab, and modernNCA are implemented based on TALENT~\cite{liu2025talent} with default hyperparameter configurations.
For Curated LLM and TabGen-ICL, we use DeepSeek v3.2 as the inference model. Raw LLM baselines are queried via their respective APIs.
To ensure fairness, all approaches share the same prompt templates and strict input constraints, with details provided in Appendix~\ref{appendix:prompt templates}.

\paragraph{Evaluation Metrics.}
For the intra-cohort setting, we report Mean Squared Error (MSE) and $R^2$ to assess overall generative accuracy and data fidelity, Median Absolute Error (MedAE) to capture typical-sample deviation, and the 95th percentile of absolute errors (P95) to evaluate robustness against extreme outliers. Higher $R^2$ values indicate better performance, whereas lower values are preferred for MSE, MedAE, and P95.

For the cross-cohort setting, we evaluate generative data fidelity using the Synthetic Data Vault (SDV) suite~\cite{patki2016sdv}, specifically reporting the Column Distribution and Inter-Column Relationship scores. These two metrics quantify how closely the generated target-domain features match the ground-truth data in terms of marginal per-feature distributions and inter-feature dependencies, respectively. Furthermore, we introduce an overall Average Score, calculated as the mean of these two SDV metrics.

\subsection{Performance Evaluation}

\begin{table}[t]
\centering
\scriptsize
\setlength{\tabcolsep}{6pt}
\renewcommand{\arraystretch}{1.15}
\caption{Performance comparison of intra-cohort feature expansion on {\PrivateDataset} (generating microbiota from metabolites). \textbf{Bold} and \underline{underlined} values denote the best and second-best results.}
\label{tab:within_table}
\begin{tabular}{llcccc}
\toprule
\multicolumn{2}{c}{Methods} & MSE($\times 10^{-4}$)$\downarrow$ & $R^2\uparrow$ & MedAE($\times 10^{-3}$)$\downarrow$ & P95($\times 10^{-2}$)$\downarrow$ \\
\midrule
\multirow{6}{*}{\shortstack{Tabular\\Model}}
& XGBoost~\cite{chen2016xgboost}                    & 10.787 & 0.561 & 7.886  & 1.469 \\
& FT-Transformer~\cite{gorishniy2021revisiting}    & 13.220 & 0.462 & 13.032 & 2.655 \\
& DANets~\cite{chen2022danets}                     & 12.222 & 0.502 & 8.446  & 1.627 \\
& SwitchTab~\cite{wu2024switchtab}                 & 12.618 & 0.486 & 15.064 & 2.840 \\
& modernNCA~\cite{ye2025revisiting}                & 14.445 & 0.412 & 8.317  & 1.700 \\
& TabPFN~\cite{hollmann2025tabpfn}                 & \underline{10.207} & \underline{0.584} & \underline{7.743} & \textbf{1.347} \\
\midrule
\multirow{4}{*}{LLM}
& Curated LLM~\cite{seedat2024curatedllm}          & 14.124 & 0.425 & 8.789 & 1.598 \\
& TabGen-ICL~\cite{fang2025tabgen}                 & 13.120 & 0.467 & 9.074 & 1.436 \\
& DeepSeek v3.2~\cite{liu2025deepseek}             & 13.627 & 0.445 & 8.171 & 1.573 \\
& Baichuan M3-plus~\cite{dou2026baichuanm3}        & 14.464 & 0.411 & 8.482 & 1.505 \\
\midrule
\multicolumn{2}{c}{\textbf{{\model}~(Ours)}} 
& \textbf{9.714} & \textbf{0.604} & \textbf{7.531} & \underline{1.378} \\
\bottomrule
\end{tabular}
\end{table}

\subsubsection{Intra-cohort Setting}
Table~\ref{tab:within_table} reports the intra-cohort feature expansion results on {\PrivateDataset}, where microbiota features are generated from metabolite inputs. The results demonstrate that {\model} achieves the best overall performance, outperforming all baselines. 

Among the tabular methods, TabPFN yields the strongest baseline performance, ranking second on both MSE and $R^2$. This indicates that tabular foundation models are competitive for intra-cohort feature expansion. However, {\model} consistently improves over TabPFN across most metrics, underscoring the necessity and benefit of injecting biomedical knowledge into tabular foundation modeling. In contrast, LLM-based baselines generally underperform compared with tabular methods, suggesting that relying solely on broad semantic and medical knowledge is insufficient for this task without effectively modeling tabular structural dependencies. 

Furthermore, {\model} achieves the lowest MedAE, indicating a higher fidelity of the generated features to the ground-truth observations on typical patient samples. Regarding P95, although {\model} is suboptimal, it remains highly competitive and close to the best-performing baseline. This confirms that {\model} successfully maintains control over high-error tail cases while securing a significant advantage on typical samples. 

To ensure comprehensive evaluation, we also conducted bidirectional inference experiments (i.e., expanding metabolite features using microbiota inputs). These results further validate the effectiveness and robustness of {\model}, with detailed findings provided in Appendix~\ref{appendix:supplementary feature expansion experiment}.

\subsubsection{Cross-cohort Setting}

\begin{table}[t]
\centering
\scriptsize
\setlength{\tabcolsep}{8pt}
\renewcommand{\arraystretch}{1.15}
\caption{Performance comparison of cross-cohort feature expansion from {\PrivateDataset} to {\PublicDataset} (generating microbiota from metabolites). \textbf{Bold} and \underline{underlined} values denote the best and second-best results. The Avg.\ Score is the mean of Column Distribution and Inter-Column Relationship.}
\label{tab:cross_table}
\begin{tabular}{lccc}
\toprule
Methods & Column Distribution $\uparrow$ & Inter-Column Relationship $\uparrow$ & Avg. Score $\uparrow$ \\
\midrule
XGBoost~\cite{chen2016xgboost} & 0.093 & 0.685 & 0.389 \\
FT-Transformer~\cite{gorishniy2021revisiting} & 0.117 & 0.672 & 0.395 \\
DANets~\cite{chen2022danets} & 0.106 & 0.702 & 0.404 \\
SwitchTab~\cite{wu2024switchtab} & \underline{0.313} & \underline{0.805} & \underline{0.559} \\
modernNCA~\cite{ye2025revisiting} & 0.025 & 0.803 & 0.414 \\
TabPFN~\cite{hollmann2025tabpfn} & 0.234 & \textbf{0.822} & 0.527 \\
\textbf{{\model}~(Ours)} & \textbf{0.432} & 0.762 & \textbf{0.597} \\
\bottomrule
\end{tabular}
\end{table}

Table~\ref{tab:cross_table} reports the cross-cohort feature expansion results. In this subsection, the entirety of {\PrivateDataset} along with the observed metabolite domain from {\PublicDataset} is used to infer the uncollected microbiota domain in {\PublicDataset}. Given the universally worse performance of LLM-based methods observed in the intra-cohort evaluation, we exclude them from this setting.

The results show that {\model} achieves the best overall generation quality with the highest Average Score of 0.597, outperforming all tabular baselines under severe distribution shifts. Specifically, {\model} obtains the best Column Distribution score of 0.432, substantially surpassing the strongest baseline, SwitchTab. This highlights that the synthesized microbiota features produced by {\model} more accurately reflect the marginal distribution of the real target dataset. Although {\model} does not achieve the absolute highest Inter-column Relationship score, it remains highly competitive. 


\subsection{Ablation Study}

    

\subsubsection{Effectiveness of Knowledge Graph}

To validate the knowledge injection mechanism, we evaluate {\model} against its backbone, \textbf{raw TabPFN}, alongside three modified graph variants: \textbf{Complete Graph}, \textbf{Random Graph}, and \textbf{Single-domain Graph}. This setup allows us to discuss the utility of external knowledge from the impacts of information volume and topological selection.

\begin{wraptable}{r}{0.5\textwidth} 
\centering
\footnotesize
\caption{Ablation study on knowledge injection mechanism of {\model} on intra-cohort setting. }
\label{tab:ablation_results}
\begin{tabular}{lcc}
\toprule
Methods & MSE($\times 10^{-4}$)$\downarrow$ & $R^2\uparrow$ \\
\midrule
\textbf{\model} & \textbf{9.714} & \textbf{0.604} \\
raw TabPFN & 10.207 & 0.584 \\
w/ Complete Graph & \underline{10.123} & \underline{0.588} \\
w/ Random Graph & 10.133 & 0.587 \\
w/ Single-domain Graph & 11.089 & 0.551 \\
\bottomrule
\end{tabular}
\end{wraptable}

Table~\ref{tab:ablation_results} shows that {\model} achieves the best performance. Compared with its base TabPFN, {\model} yields significant improvements, demonstrating the overall efficacy of incorporating curated biomedical priors. Second, the performance degradation in both the complete and random graph settings confirms that the superiority of the full {\model} stems from meaningful biomedical structures rather than volume. Most notably, the single-domain graph variant suffers the most severe performance drop, performing even worse than the raw TabPFN. This phenomenon indicates that injecting only single-domain external information misguides the representation, causing the model to overfit to local intra-domain noise. It strongly underscores that explicitly modeling inter-domain biomedical associations is the critical key to preventing negative transfer and enabling robust cross-domain feature expansion.

\subsubsection{Robustness to Incomplete Knowledge Graphs}

In practice, curated biomedical knowledge graphs may provide incomplete coverage for rare diseases or under-studied populations. To simulate this scenario, we evaluate {\model}'s robustness by randomly dropping 5\%, 10\%, 20\%, and 50\% of nodes or edges before constructing the biomedical knowledge injection.


\begin{wrapfigure}{r}{0.55\linewidth} 
    \centering
    \includegraphics[width=0.9\linewidth]{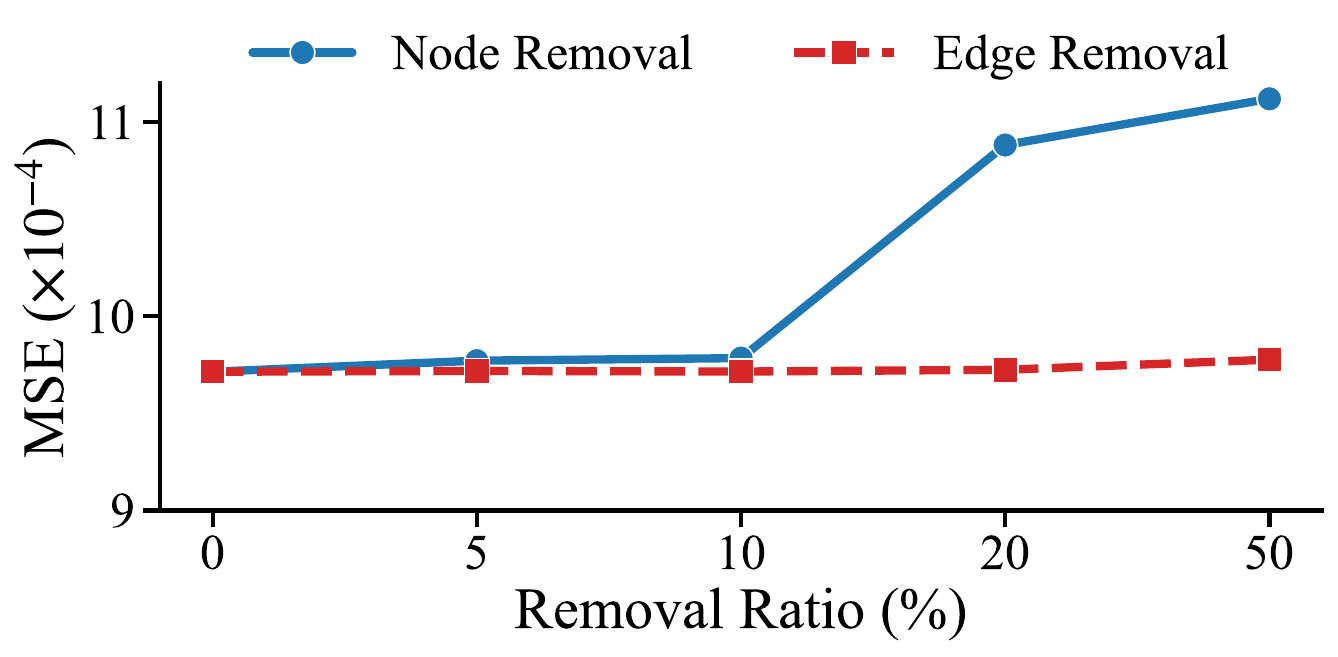}
    \caption{Robustness of {\model} to incomplete knowledge graphs on intra-cohort setting.}
    \label{fig:Robustness}
\end{wrapfigure}

As illustrated in Fig.~\ref{fig:Robustness}, edge removal induces a negligible impact, with the MSE remaining relatively flat (around $9.71 \sim 9.78 \times 10^{-4}$) across all dropout ratios. Conversely, while node removal is stable at low dropout ratios (5\%–10\%), it exhibits a pronounced degradation in data fidelity from 20\% dropout ratio, where the predictive error reaches its maximum, with the MSE peaking at $11.12 \times 10^{-4}$. This phenomenon suggests that {\model} remains effective even with an incomplete knowledge graph. Maintaining the direct mapping between tabular features and graph nodes is fundamentally more important than preserving every internal relational edge.

\subsection{Hyperparameter Sensitivity Analysis}

\begin{wrapfigure}{r}{0.55\linewidth} 
\centering
\includegraphics[width=1\linewidth]{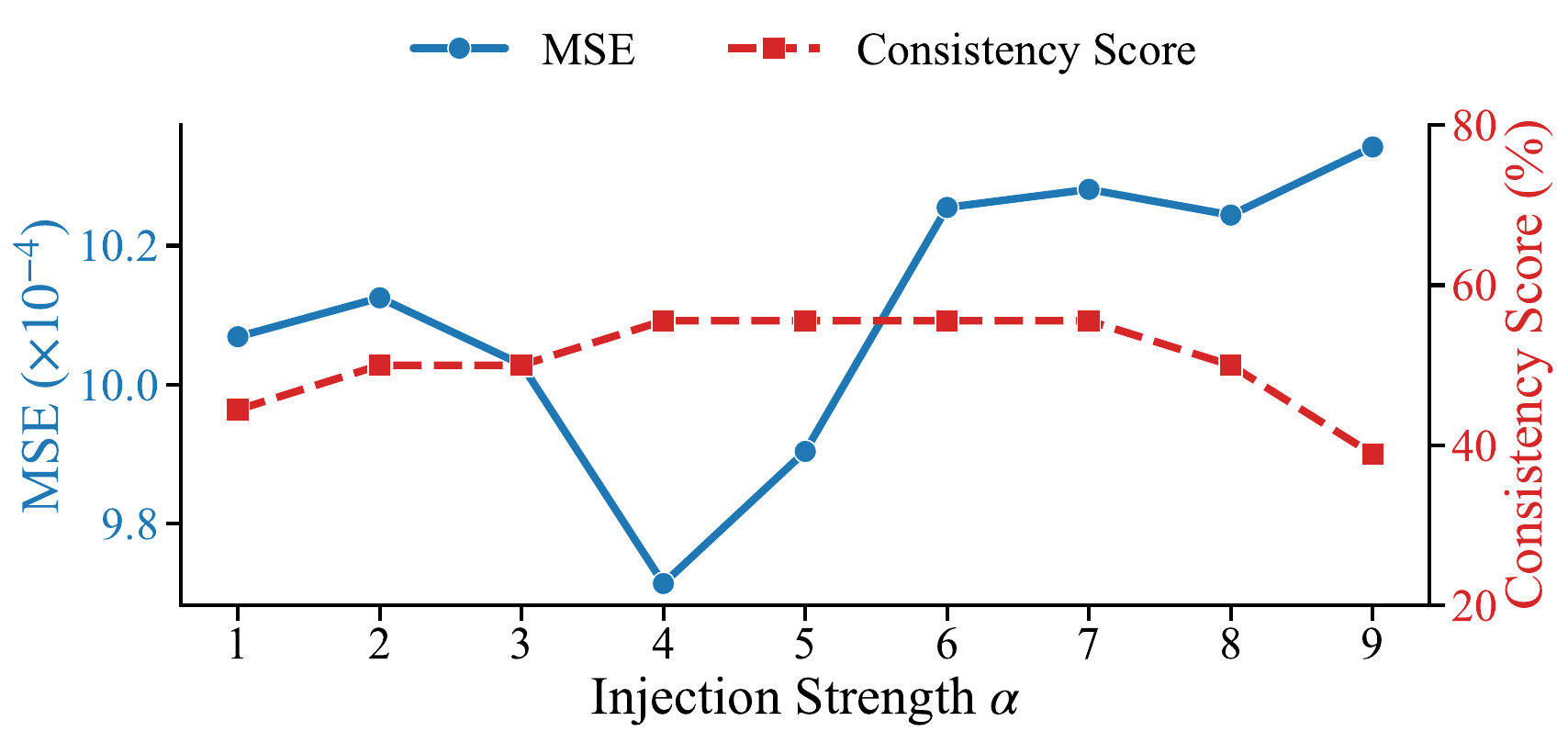}
\caption{Sensitivity of {\model} to the graph injection strength $\alpha$ on the intra-cohort setting. }
\label{fig:hyperparameter_alpha}
\end{wrapfigure}

We analyze the sensitivity of {\model} to the parameter $\alpha$, which explicitly controls the proportion of the external biomedical prior injected into the tabular data representations. As Fig.~\ref{fig:hyperparameter_alpha} shows, {\model} achieves the lowest MSE on the intra-cohort setting at $\alpha = 4$. Settings of $\alpha$ below 4 result in an insufficient injection of knowledge guidance, whereas larger values degrade performance, as excessive external knowledge overrides the useful information from the data channel.

To understand this phenomenon, we investigate the alignment between the biomedical prior from the knowledge graph and the learned information from the data channel. For each TabPFN layer, we compute the cosine similarity between the normalized knowledge matrix $\tilde{A}_w$ and the corresponding feature attention matrix. This metric quantifies the structural agreement between the injected external knowledge and the intrinsic data-driven feature dependencies. We define a consistency score as the proportion of layers with non-negative cosine similarity. This score acts as an indicator of network-wide agreement: higher values mean the injected graph more closely matches the data-driven attention patterns.
Fig.~\ref{fig:hyperparameter_alpha} shows that the consistency score initially rises but then declines, reaching its peak at $\alpha = 4$. This demonstrates that while moderate injection helps provide meaningful relationships, excessive external knowledge could conflict with intrinsic data patterns. $\alpha = 4$ achieves an optimal synergy, providing essential biomedical guidance without overriding core information from the data channel. A more detailed parameter analysis is provided in Appendix~\ref{appendix:Effect of Knowledge Graph Injection on Transformer Attention}.

%% file: 5-Conclusion.tex
\section{Conclusion}
\label{sec:conclusion}

We present {\model}, a knowledge-injected framework for cross-domain medical feature expansion. Unlike previous methods that inject superficial knowledge information, {\model} achieves an optimal synergy between the data and knowledge channels. It models inter-domain biomedical associations, providing vital biomedical guidance without overriding the intrinsic data representations. Built upon the TabPFN, {\model} operates directly on raw structured tabular data, inherently capturing numerical distributions and fundamentally outperforming LLMs that fail to preserve tabular structures and dependencies. Extensive experiments validate {\model}'s superiority and robustness in both intra- and cross-cohort settings.

%% file: 6-Appendix.tex
\section{Appendix}
\subsection{Prompt Templates for LLM-based Baselines}
\label{appendix:prompt templates}

To facilitate reproducibility, we detail the prompt templates utilized for the LLM-based baselines. As discussed in the main text, we compare {\model} against four LLM-based approaches: CLLM (Curated LLM), TabGen-ICL, DeepSeek v3.2, and Baichuan-M3-plus. Notably, both CLLM and TabGen-ICL employ DeepSeek v3.2 as their backbone model. To ensure a fair evaluation, we apply a unified system and user prompt templates across all LLM baselines. 

The system prompt used in our experiments is presented below.

\begin{promptbox}{System Prompt}
You are a synthetic data generator tasked with completing metabolite-only samples into full samples.

1. Analyze the provided real samples carefully. Each real sample is a JSON object containing BOTH metabolite fields and microbiota fields.
2. When given metabolite-only samples, infer and generate the missing microbiota fields based on patterns learned from the real samples.
3. Maintain realistic relationships/correlations between metabolite and microbiota fields as reflected in the real samples.

IMPORTANT (must-follow):
- For each metabolite-only sample, you MUST copy the metabolite fields and their values EXACTLY as provided.
  Do NOT change, normalize, round, reorder, rename, or regenerate metabolite values.
- ONLY generate the missing microbiota fields so that the completed sample has the SAME set of keys as real samples.
- Do NOT use a trivial completion such as filling all microbiota fields with 0 or the same constant template,
  unless this is strongly supported by very similar real samples.
- Microbiota feature values must reflect the real-sample distributions: use realistic ranges, non-zero rates, and variability seen in real samples.
  If a microbiota feature is frequently non-zero in real samples, it should not be set to 0 by default.

Output (STRICT):
- Output ONLY a valid JSON array, with length equal to the number of metabolite-only input lines, in the same order.
- Do NOT wrap in Markdown/code fences.
- Do NOT output any extra text before or after the JSON.
\end{promptbox}

The user prompt template is shown below.

\begin{promptbox}{User Prompt}
Real Samples (each line is one JSON object):
<<<REAL_SAMPLES_START>>>
{real_samples}
<<<REAL_SAMPLES_END>>>

Metabolite-only Samples to Complete (each line is one JSON object):
<<<METABOLITE_ONLY_START>>>
{metabolite_only_samples}
<<<METABOLITE_ONLY_END>>>
\end{promptbox}

These templates ensure that all LLM-based baselines receive the same input context and output constraints, thereby improving the reproducibility and fairness of the comparison.
For the reverse-direction supplementary experiment in Appendix~\ref{appendix:supplementary feature expansion experiment}, we use the same templates, with metabolite and microbiota fields swapped accordingly.

\subsection{Reverse Feature Expansion: From Microbiota to Metabolites}
\label{appendix:supplementary feature expansion experiment}

\begin{table}[t]
\centering
\small
\setlength{\tabcolsep}{6pt}
\renewcommand{\arraystretch}{1.15}
\caption{Performance evaluation of cross-domain feature expansion methods from microbiota to metabolites on {\PrivateDataset}. Lower MSE, MedAE, and P95 indicate better performance, whereas higher $R^2$ indicates better performance. \textbf{Bold} and \underline{underlined} numbers denote the best and second-best results, respectively. $\uparrow$ and $\uparrow(\%)$ report the absolute and relative improvements of {\model} over the second-best baseline when {\model} achieves the best result.}
\label{tab:supplementary_within_table}
\begin{tabular}{llcccc}
\toprule
\multicolumn{2}{c}{Methods} 
& MSE($\times 10^2$)$\downarrow$ 
& $R^2\uparrow$ 
& MedAE$\downarrow$ 
& P95$\downarrow$ \\
\midrule
\multirow{6}{*}{\shortstack{Tabular\\Model}}
& XGBoost~\cite{chen2016xgboost}                 
& \underline{9.767}  
& \underline{0.415} 
& 111.857 
& \textbf{318.271} \\
& FT-Transformer~\cite{gorishniy2021revisiting} 
& 16.901             
& -0.013            
& 80.938  
& 510.080 \\
& DANets~\cite{chen2022danets}                  
& 23.102             
& -0.384            
& 119.033 
& 507.069 \\
& SwitchTab~\cite{wu2024switchtab}              
& 16.861             
& -0.010            
& \textbf{77.891} 
& 506.462 \\
& modernNCA~\cite{ye2025revisiting}             
& 10.010             
& 0.400             
& 114.589 
& 351.722 \\
& TabPFN~\cite{hollmann2025tabpfn}              
& 10.041             
& 0.398             
& \underline{80.697} 
& 369.804 \\
\midrule
\multirow{4}{*}{LLM}
& Curated LLM~\cite{seedat2024curatedllm}       
& 11.606             
& 0.304             
& 108.022 
& 370.197 \\
& TabGen-ICL~\cite{fang2025tabgen}              
& 12.666             
& 0.241             
& 84.398  
& 404.260 \\
& DeepSeek v3.2~\cite{liu2025deepseek}          
& 12.734             
& 0.237             
& 89.224  
& 393.801 \\
& Baichuan M3-plus~\cite{dou2026baichuanm3}     
& 13.405             
& 0.197             
& 141.518 
& 392.842 \\
\midrule
\multicolumn{2}{c}{\textbf{{\model} (Ours)}} 
& \textbf{9.189} 
& \textbf{0.449} 
& 100.706 
& \underline{319.612} \\
\multicolumn{2}{c}{$\uparrow$} 
& 0.578 
& 0.034 
& / 
& / \\
\multicolumn{2}{c}{$\uparrow(\%)$} 
& 5.92 
& 8.19 
& / 
& / \\
\bottomrule
\end{tabular}
\end{table}

In the main paper, we evaluate all methods on the task of cross-domain feature expansion from metabolites to microbiota in {\PrivateDataset}.
As a supplementary experiment, we further evaluate the reverse-direction task, namely expanding metabolite features from microbiota features. 
The experiment is intended to test whether the proposed method remains effective when the source and target domains are swapped.

We strictly follow the same experimental setting as in the main intra-cohort setting, including the data split, baselines, and evaluation metrics. 
In particular, the main paper uses a 40/60 collected-uncollected split on {\PrivateDataset}, retains the top 50 features from each domain, and evaluates intra-cohort regression performance using MSE and $R^2$, while also reporting MedAE and P95 in the main results table.

Table~\ref{tab:supplementary_within_table} reports the results of the supplementary expansion task. 
{\model} achieves the best performance on MSE and $R^2$, indicating the strongest overall expansion quality among baselines. 
It also remains competitive on P95, suggesting relatively strong robustness on hard samples in the high-error tail.
On MedAE, however, {\model} does not rank first, which indicates that there is still room for improvement on the typical-sample error level. 

\begin{figure}
    \centering
    \includegraphics[width=0.6\linewidth]{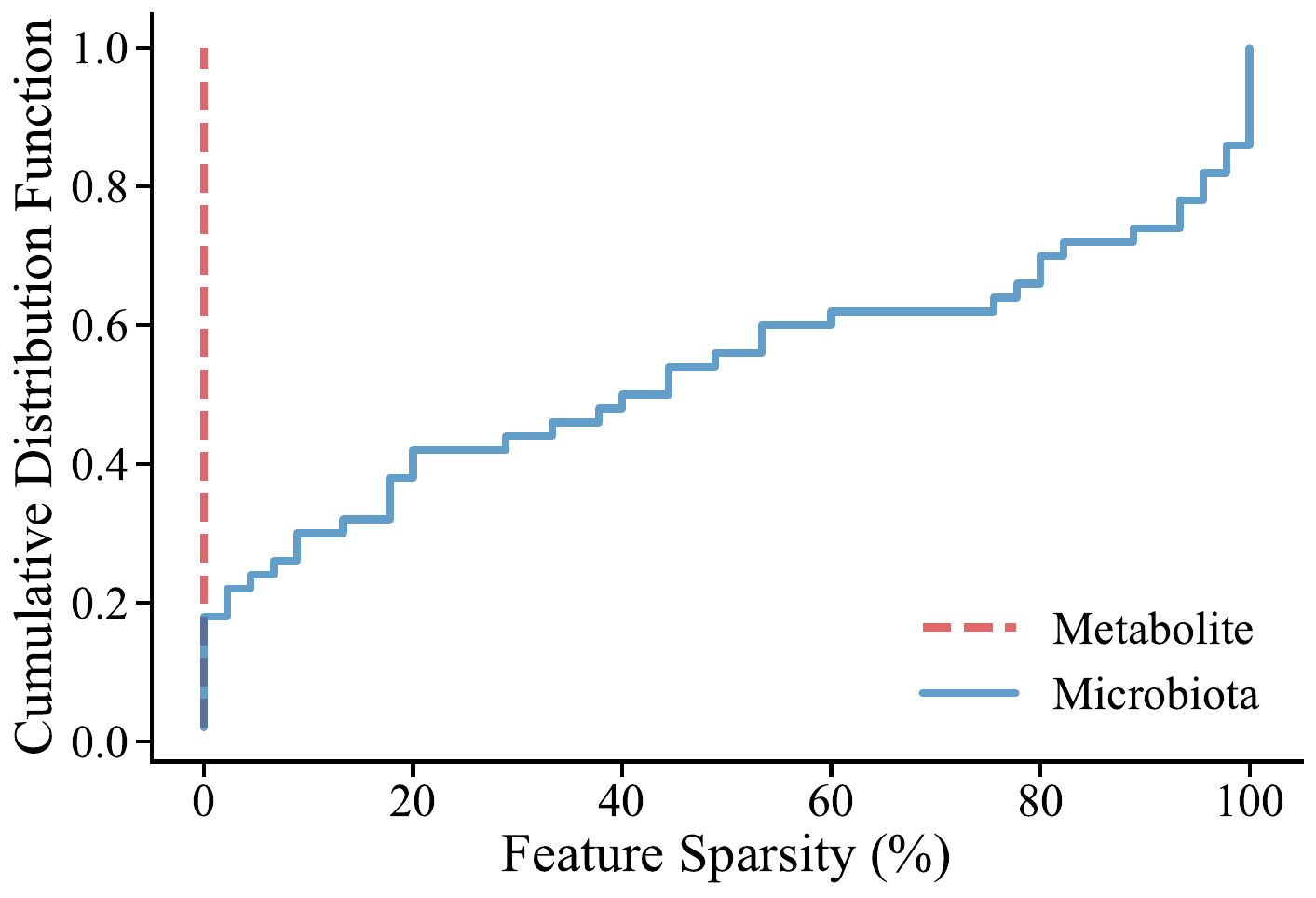}
    \caption{Feature-wise sparsity distributions of metabolite and microbiota features on {\PrivateDataset}. Sparsity is measured as the proportion of zero values in each feature. The cumulative distributions show that microbiota features are substantially sparser than metabolite features, which helps explain the difficulty of the feature expansion task from microbiota to metabolites.}
    \label{fig:Zero_Ratio_CDF}
\end{figure}

Notably, compared with the original metabolite-to-microbiota task, all methods perform worse on this reverse  microbiota-to-metabolites feature expansion task.
This degradation stems from the inherent sparsity of microbiota data, which provides a fundamentally weaker source-domain signal. As illustrated in Fig.~\ref{fig:Zero_Ratio_CDF}, 19 microbiota features exhibit zero values in over 50\% of patients, a high sparsity level entirely absent in the selected metabolite features. This severe zero-inflation substantially reduces the effective information available for learning reliable cross-domain dependencies.

Despite this inherent sparsity challenge, {\model} consistently achieves the best overall MSE and $R^2$. These results confirm that our knowledge-injected framework effectively generalizes across varying biomedical feature-expansion directions.

\begin{table}[t]
\centering
\caption{Downstream evaluation on the disease detection task using raw versus expanded training data. Balanced Accuracy, F1, and AUC are reported for Random Forest, MLP, and XGBoost. \textbf{Bold} and \underline{underlined} numbers denote the best and second-best results, respectively.}
\label{tab:downstream_eval}
\begin{tabular}{l l c c c}
\toprule
Model & Trainset & Balanced ACC & F1 & AUC \\
\midrule
\multirow{2}{*}{Random Forest}
& Generated & \textbf{0.863} & \underline{0.863} & \textbf{0.943} \\
& Raw  & \underline{0.776} & \textbf{0.871} & \underline{0.909} \\
\midrule
\multirow{2}{*}{MLP}
& Generated & 0.741 & 0.833 & 0.798 \\
& Raw  & 0.764 & 0.821 & 0.807 \\
\midrule
\multirow{2}{*}{XGBoost}
& Generated & 0.732 & 0.634 & 0.732 \\
& Raw  & 0.618 & 0.812 & 0.873 \\
\bottomrule
\end{tabular}
\end{table}

\subsection{Downstream Utility of Expanded Features}

We conduct a downstream classification experiment to further evaluate whether the expanded features are functionally comparable to the real features.
Specifically, we consider a disease detection task, in which metabolite features and microbiota features are used to determine the health status of each subject.
We split the original dataset into 60\% collected data and 40\% held-out uncollected data. 
For the collected data, we construct an expanded dataset by replacing the target-domain features with the generated ones, while keeping the real disease labels unchanged. 
The held-out 40\% original data are used only for testing.
We evaluate three representative and architecture-neutral classifiers, including Random Forest, MLP, and XGBoost, to avoid introducing evaluation bias from additional domain-specific modules such as graph-based structures or knowledge-injection components.
Balanced Accuracy, F1 score, and AUC are adopted as evaluation metrics to provide a comprehensive assessment of classification performance.

The results are shown in Table~\ref{tab:downstream_eval}.
Overall, classifiers trained on expanded data achieve performance broadly comparable to those trained on original data, indicating that the expanded data preserve the key discriminative information required for disease detection.
Moreover, the overall relative performance pattern across classifiers remains largely unchanged under the two training settings, suggesting that the expanded data do not disproportionately favor any particular model.
For MLP, the results under the two training settings are very close across all three metrics, further supporting the functional consistency between expanded and original data.
Overall, the downstream evaluation suggests that the expanded features are not merely numerically close to the original features, but also preserve disease-relevant discriminative information for practical classification tasks.

\subsection{Effect of Knowledge Graph Injection Strength}
\label{appendix:Effect of Knowledge Graph Injection on Transformer Attention}
\begin{figure}
    \centering
    \includegraphics[width=0.7\linewidth]{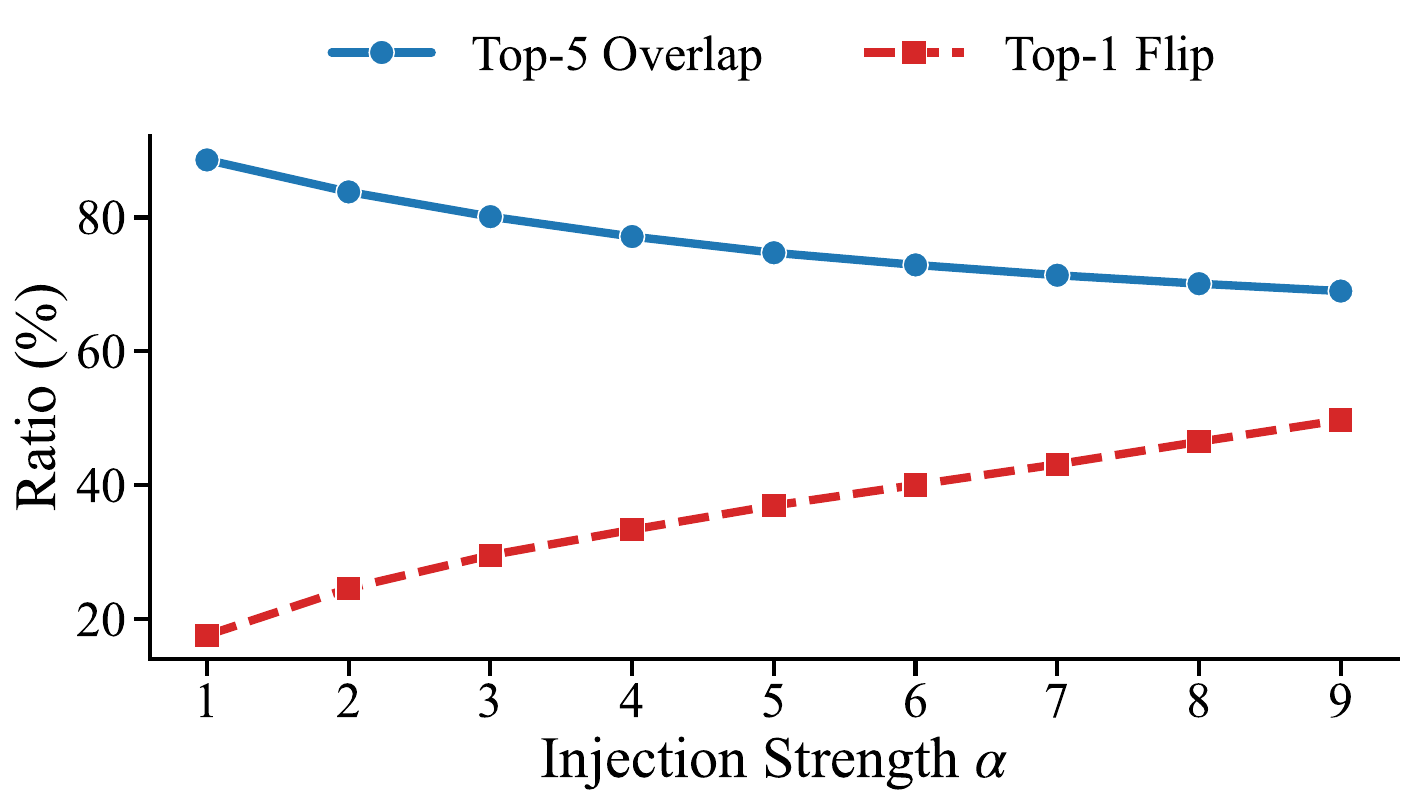}
    \caption{Effect of knowledge graph injection on the native attention ranking of the TabPFN backbone. Top-5 Overlap measures the consistency between the top-5 attended features before and after graph injection, while Top-1 Flip measures how often the top-1 attended feature changes; larger injection strength leads to lower overlap and higher flip ratio, indicating stronger rewriting of the original attention organization.}
    \label{fig:RankPerturbationMetrics}
\end{figure}

The main paper shows that the performance of {\model} is sensitive to the injection strength $\alpha$.
Specifically, the best performance is achieved at $\alpha$ = 4, whereas overly large $\alpha$ decreases the consistency score and worsens MSE, indicating that excessively strong graph injection interferes with the model’s original data-driven attention structure.
To better understand the phenomenon, we further analyze how knowledge graph injection changes the native attention ranking of the TabPFN backbone.

We use two ranking-based metrics in the following analysis. 
\texttt{Top-5 Overlap} measures the overlap between the top-5 attended features before and after graph injection, where a higher value indicates greater consistency with the original attention ranking. 
\texttt{Top-1 Flip} measures how often the top-1 attended feature in each row changes after graph injection, where a higher value indicates more frequent rewriting of the original attention target.

Results in Fig.~\ref{fig:RankPerturbationMetrics} show a clear trend: as the injection strength increases, the enhanced attention increasingly departs from the original attention ranking. 
Specifically, \texttt{Top-5 Overlap} decreases from 89.13\% at $\alpha=1$ to 77.63\% at $\alpha=4$ and 69.93\% at $\alpha=9$, while \texttt{Top-1 Flip} increases from 17.94\% to 34.93\% and 55.29\%, respectively.
When the injection strength is relatively small, the knowledge graph prior mainly acts as auxiliary structural guidance and preserves most of the native attention preference. 
When the injection strength becomes large, however, the knowledge graph prior no longer provides only supplementary information; instead, it begins to rewrite the original attention organization over a broad range of rows. 
Therefore, the best injection strength should strike a balance between incorporating external biomedical guidance and preserving useful native dependencies learned from data.

\subsection{Case Study: Attention Re-ranking via KG Injection}

To elucidate how KG injection modulates feature preferences, we analyzed attention-score re-ranking for the \textit{Prevotellaceae} family. Following our intra-cohort setting, metabolite source features were partitioned into 17 groups (3 per group) to infer the target feature. We computed average attention scores across all layers and heads before and after injection; a higher rank shift indicates an increased contribution to feature fusion.

\begin{figure}
    \centering
    \includegraphics[width=0.7\linewidth]{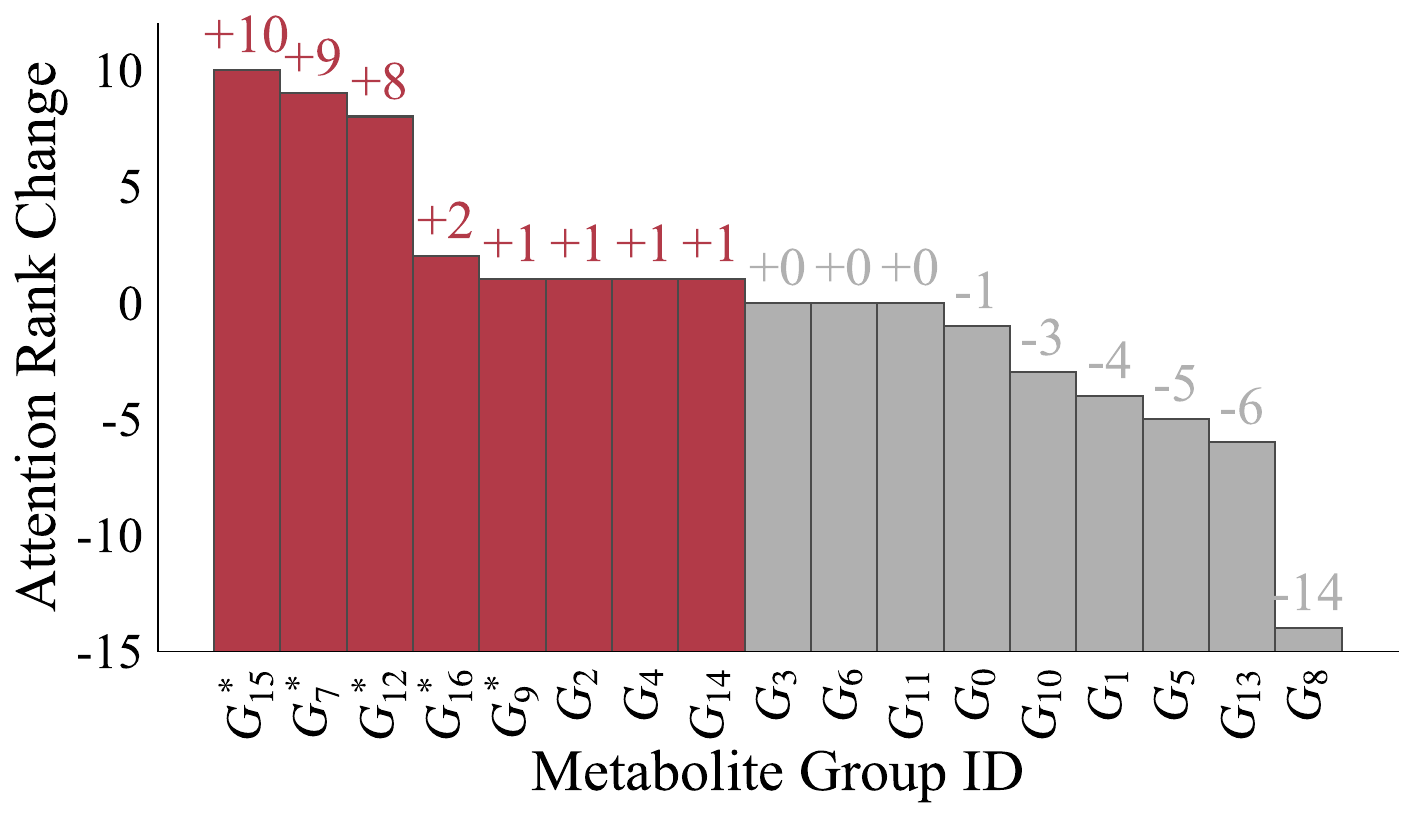}
    \caption{Case study of attention-rank changes in metabolite feature groups after knowledge graph injection. Bar tops indicate rank changes after injection. G$_i$ denotes the $i$-th metabolite feature group, and $*$ marks groups containing metabolite features in SPOKE. Red and gray bars indicate increased and non-increased ranks, respectively.}
    \label{fig:CaseStudy}
\end{figure}

As shown in Fig. ~\ref{fig:CaseStudy}, KG injection systematically prioritizes biologically relevant metabolites. Groups mapped to the SPOKE KG exhibited the most significant upward shifts, notably $G_{15}^*$ (Picolinic acid, +10), $G_7^*$ (Tyrosine, +9), and $G_{12}^*$ (Leucine, +8). Other KG-informed groups like $G_{16}^*$ (N-Acetylglucosamine, +2) and $G_9^*$ (Valine, +1) also displayed positive shifts, whereas unmapped groups generally declined or remained stable.

This prioritization aligns with empirical evidence. \textit{Prevotellaceae} strains are key participants in amino-acid metabolism, particularly involving branched-chain amino acids like Leucine and Valine~\cite{Atasoglu1998PrevotellaAminoAcids,DeFilippis2019PrevotellaDiet}. Furthermore, Prevotellaceae-mediated mucin degradation involves enzymes acting on N-acetylglucosamine substrates~\cite{Wright2000MucinSulfatase,Wright2000PrevotellaMucin}, while the biomedical database KEGG~\cite{Kanehisa2000KEGG} annotations confirm aromatic amino acid metabolism modules for Tyrosine. The promotion of Picolinic acid further suggests alignment with tryptophan-related pathways often altered in gut dysbiosis.

Consistent with Appendix~\ref{appendix:Effect of Knowledge Graph Injection on Transformer Attention}, these results demonstrate that KG priors provide targeted guidance rather than uniform bias. By reshaping attention toward biologically grounded features, MedKGTab fundamentally enhances model interpretability.

\subsection{Limitations}
\label{appendix:limitations}

While MedKGTab demonstrates promising performance in cross-domain feature expansion, we acknowledge two primary limitations regarding its knowledge-injection mechanism. 
\begin{itemize}
  \item Its generalizability is bottlenecked by the semantic alignment between clinical variables and knowledge graph nodes. Since real-world tabular data often contain non-standard abbreviations, automated mapping risks misalignment noise and negative transfer without expert curation. 
  \item The framework is constrained by out-of-vocabulary features due to the inherent incompleteness of existing KGs. For highly specialized or newly discovered markers absent from the graph, MedKGTab cannot derive effective structural priors, limiting its full potential on cutting-edge datasets
\end{itemize}

Future work will focus on addressing these challenges through LLM-based semantic matching modules and dynamic graph expansion via literature knowledge mining.